\documentclass[11pt]{article}
\usepackage[utf8]{inputenc}
\usepackage{newunicodechar}
\newunicodechar{、}{,}
\usepackage[preprint]{acl}

\usepackage{times}
\usepackage{latexsym}
\usepackage{booktabs}
\usepackage{siunitx}
\usepackage{multirow}
\usepackage[table]{xcolor}
\usepackage{caption}
\usepackage[normalem]{ulem}
\usepackage{array}
\usepackage{tcolorbox}
\tcbuselibrary{breakable}
\usepackage{bbm}
\usepackage{amssymb}
\usepackage{algorithm}
\usepackage{algpseudocode}

\usepackage{amsmath}
\usepackage{geometry}
\usepackage[utf8]{inputenc}
\usepackage{newunicodechar}
\newunicodechar{（}{(}
\newunicodechar{）}{)}

\usepackage[T1]{fontenc}
\usepackage{amsfonts}

\usepackage[utf8]{inputenc}

\usepackage{microtype}

\usepackage{inconsolata}

\usepackage{graphicx}

\title{PRISM: Prompt Refinement via Image-grounded Self-rewarding Mechanism for Text-to-Image Generation}

\author{
 \textbf{Guo Tang\textsuperscript{1}},
 \textbf{HongJie Luo\textsuperscript{2}},
 \textbf{Tianxu Wang\textsuperscript{3}},
 \textbf{Ying Zhang\textsuperscript{4}},
 \textbf{Hao Wang\textsuperscript{3}},
\\
\\
 \textsuperscript{1}Harbin Institute of Technology, Shenzhen,
 \textsuperscript{2}Sun Yat-sen University,
 \\
 \textsuperscript{3}South China University of Technology,
 \textsuperscript{4}Guangdong University of Technology
\\
 \small{
   \textbf{Correspondence:} \href{mailto:email@domain}{guot3907@gamil.com}
 }
}

\begin{document}
\maketitle
\begin{abstract}
Text-to-image generation models can synthesize high-quality images
from natural language descriptions, but their performance remains highly
sensitive to prompt formulation. Existing prompt optimization methods
mainly rely on text-side rewriting, prompt expansion, or external reward
signals, offering limited image-grounded diagnosis and weak support for
learning reusable optimization policies.
In this paper, we propose
\textbf{PRISM}, a \textbf{P}rompt \textbf{R}efinement framework via
\textbf{I}mage-grounded \textbf{S}elf-rewarding \textbf{M}echanism.
PRISM closes the prompt-image-feedback loop by interpreting generated
images with structured visual diagnosis and scoring them along semantic
consistency, aesthetic quality, and human preference alignment. It first
initializes a unified VLM through multi-task supervised fine-tuning, and
then improves the prompt policy via self-rewarding optimization with a
hybrid ideal-point and Chebyshev reward. Extensive experiments show that
PRISM improves holistic image quality and fine-grained semantic alignment,
while providing interpretable feedback for targeted prompt refinement. The code is available at \url{https://anonymous.4open.science/r/PRISM-FF81}
\end{abstract}

\section{Introduction}

Text-to-image (T2I) generation has made remarkable progress in synthesizing high-quality images from natural language descriptions~\cite{esser2024scaling,podell2024sdxl,chen2024pixart}. 
However, its performance remains highly sensitive to prompt formulation, as under-specified prompts often fail to provide the object, attribute, style, and composition details required for controllable and faithful generation~\cite{huang2023t2i,mahajan2024prompting}.
As a result, prompt optimization has become an important technique for improving generation quality~\cite{chen2025t2i}.
\begin{figure}[t]
  \includegraphics[width=\columnwidth]{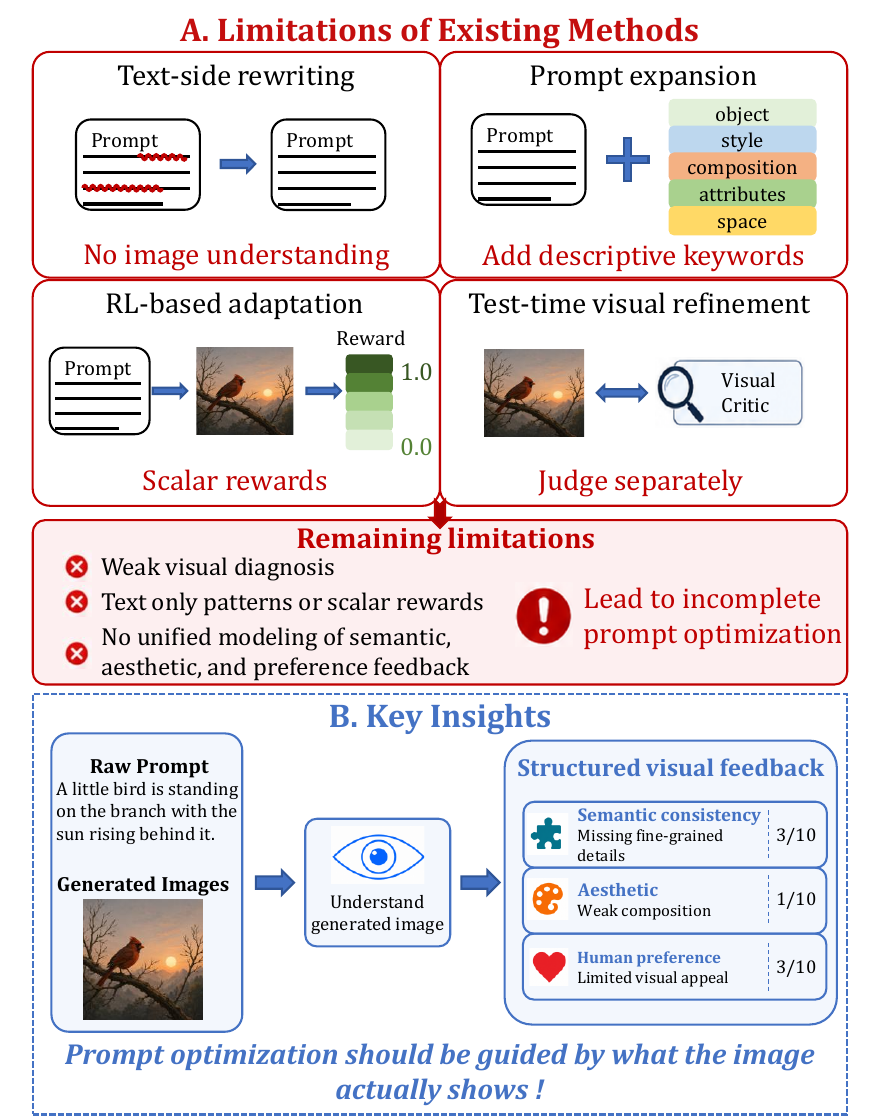}
  \caption{Motivation of the proposed PRISM framework.}
  \label{figs:motivation}
\end{figure}

Existing prompt optimization methods have explored text-side rewriting, prompt expansion, and reinforcement learning-based prompt adaptation~\cite{cao2023beautifulprompt,pavlichenko2023best,yang2025self}. 
While effective, most of them rely on textual patterns or scalar rewards, providing limited diagnostic guidance on why a generated image succeeds or fails. 
Recent visual-feedback-based methods incorporate generated images into prompt refinement, but they are often implemented as test-time pipelines with separate critic and rewriter modules, which limits their ability to learn a reusable optimization policy~\cite{wuvisualprompter,khan2025test}. 
Moreover, existing feedback signals usually emphasize isolated objectives, rather than jointly modeling semantic consistency, aesthetic quality, and preference alignment within a unified framework~\cite{zhang2024learning}.
As illustrated in Fig.~\ref{figs:motivation}, these limitations suggest that prompt optimization should be guided by structured visual feedback derived from what the generated image actually shows.

Motivated by this observation, we propose \textbf{PRISM}, a Prompt Refinement framework via Image-grounded Self-rewarding Mechanism for T2I prompt optimization.
Instead of learning prompt rewriting solely from text pairs, our method directly interprets generated images through structured visual diagnosis and uses the feedback to guide prompt optimization. 
Specifically, we train a Vision-Language Model (VLM) to assess generated images along three complementary dimensions: semantic consistency, aesthetic quality, and human preference alignment. 
The predicted diagnostic feedback and dimension-specific scores are then used for both prompt revision and reward computation. 
We initialize the VLM through multi-task supervised fine-tuning for joint prompt rewriting and visual feedback assessment, and further optimize the prompt policy via self-rewarding learning using Group Sequence Policy Optimization (GSPO)~\cite{zheng2025group}. 
A hybrid ideal-point and Chebyshev reward encourages the optimized prompts to approach the ideal visual feedback scores while penalizing weak dimensions, leading to more balanced improvements.


The main contributions of this work are summarized as follows:
\begin{itemize}
    \item \textbf{PRISM framework.}
    We propose a unified multimodal framework that learns a reusable prompt optimization policy by integrating visual feedback assessment and prompt rewriting.

    \item \textbf{A diagnosis-guided reward mechanism.}
    We introduce a multi-dimensional reward mechanism based on structured visual diagnosis, where semantic consistency, aesthetic quality, and human preference alignment are jointly considered through a hybrid ideal-point and Chebyshev objective.

    \item \textbf{Comprehensive empirical validation.}
    Extensive experiments on multiple benchmarks demonstrate the effectiveness of PRISM in prompt optimization and targeted feedback-guided refinement.
\end{itemize}

\section{Related Work}
\subsection{Text-to-Image generation}
T2I generation aims to synthesize visually realistic and semantically aligned images from natural language descriptions. 
Early studies explored GAN-based text-conditioned generation~\cite{reed2016generative,zhang2017stackgan,xu2018attngan}, while autoregressive models further formulated image synthesis as sequence modeling over visual tokens~\cite{ramesh2021zero,ding2021cogview}. 
More recently, diffusion models have become the dominant paradigm for high-fidelity T2I generation~\cite{ho2020denoising,dhariwal2021diffusion}. Representative systems such as GLIDE~\cite{nichol2021glide}, Imagen~\cite{saharia2022photorealistic}, Latent Diffusion Models~\cite{rombach2022high}, SDXL~\cite{podell2024sdxl}, and Stable Diffusion 3~\cite{esser2024scaling} have substantially improved photorealism, image-text alignment, efficiency, and high-resolution synthesis.

Despite these advances, T2I models remain highly sensitive to prompt formulation. 
Well-designed prompts are often required to specify objects, attributes, relations, style, and composition, making prompt engineering and automatic prompt optimization important directions for controllable and preference-aligned generation.
\subsection{Prompt Engineering}
Prompt engineering aims to improve the controllability, fidelity, and visual quality of T2I generation by refining input prompts. 
For example, BestPrompt searches effective prompt keywords with genetic algorithms~\cite{pavlichenko2023best}, and PEZ optimizes discrete hard prompts via gradient-based search~\cite{wen2023hard}. 
Promptist learns to adapt user inputs into model-preferred prompts using supervised fine-tuning and reinforcement learning~\cite{hao2023optimizing}, while BeautifulPrompt further leverages visual AI feedback~\cite{cao2023beautifulprompt}.
More recently, self-rewarding VLMs have been explored to unify prompt rewriting and image evaluation within a single model~\cite{yang2025self}. 
Although these methods improve prompt quality, they provide limited image-grounded diagnosis of generation failures.

Beyond text-side optimization, recent methods incorporate generated images or external evaluators into prompt refinement. 
PromptMagician supports interactive prompt exploration by retrieving similar prompt-image pairs and recommending important keywords~\cite{feng2023promptmagician}. 
OPT2I iteratively revises prompts with an LLM to maximize prompt-image consistency~\cite{manas2024improving}, while T2I-Copilot~\cite{chen2025t2i} and PromptSculptor~\cite{xiang2025promptsculptor} employ multi-agent systems for evaluator-guided refinement. 
TIR performs closed-loop test-time refinement by using an MLLM to check prompt-image alignment and update prompts with history-aware feedback~\cite{khan2025test}. 
VisualPrompter further decomposes prompts into atomic semantic concepts with Davidsonian Scene Graphs~\cite{cho2024davidsonian} and uses visual feedback to identify missing concepts for target-specific optimization~\cite{wuvisualprompter}. 
These feedback-aware methods show the value of generated images for prompt refinement, but they are often constrained by fragmented feedback objectives, reliance on external or test-time evaluators, and limited ability to transform structured visual feedback into a reusable self-optimizing prompt policy.

\section{Methodology}
\begin{figure*}[t]
  \includegraphics[width=1.0\linewidth]{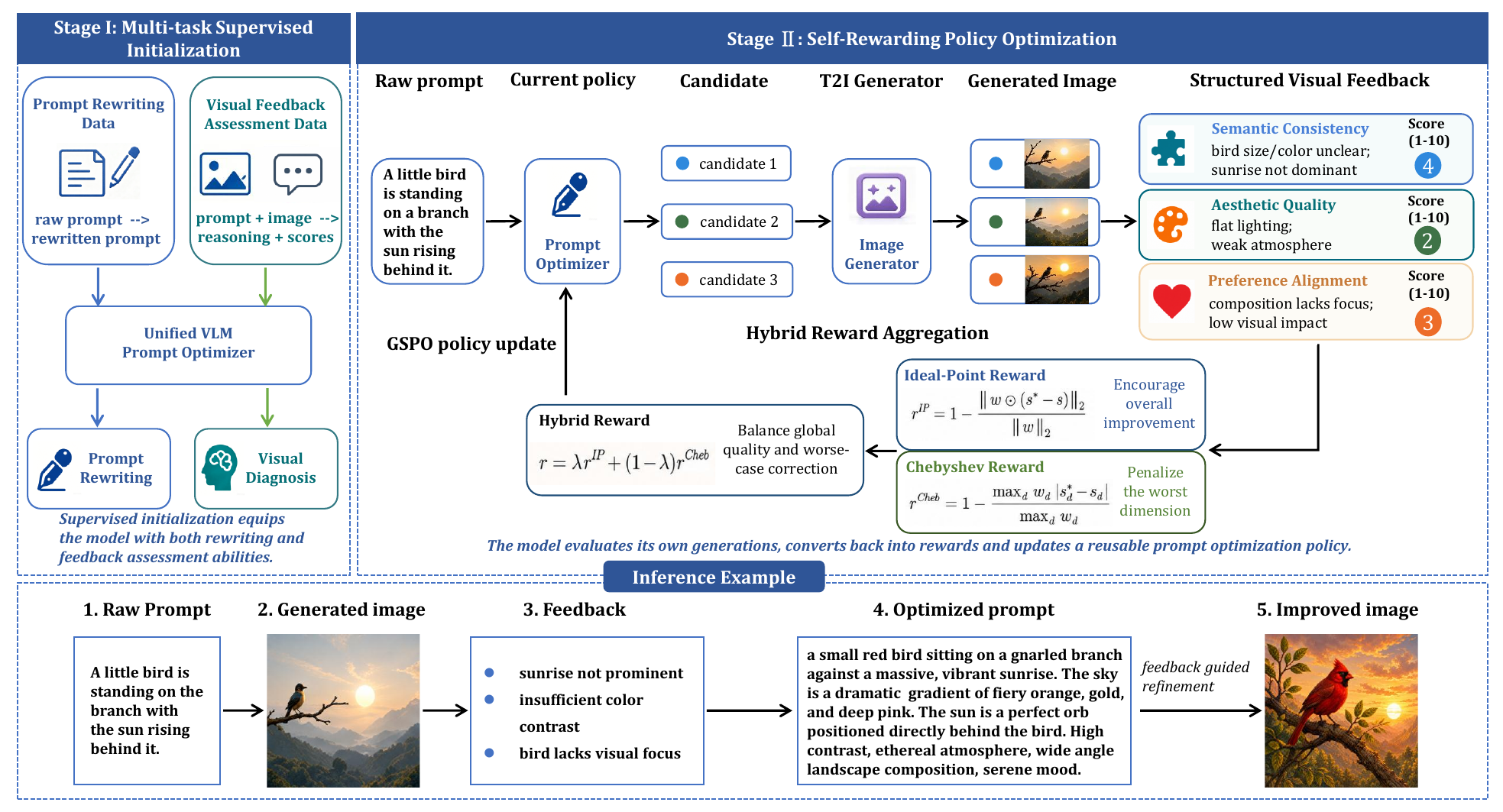}
  \caption {Overview of the proposed PRISM framework. 
  Stage I performs multi-task supervised initialization to equip the VLM with prompt rewriting and visual feedback assessment abilities. 
  Stage II optimizes the prompt policy through self-rewarding learning, where candidate prompts are rendered by a frozen T2I generator, evaluated with structured visual feedback, and updated using hybrid reward aggregation. 
  The bottom row illustrates an inference example of feedback-guided prompt refinement.}
  \label{fig:overview}
\end{figure*}

In this section, we introduce PRISM for image-grounded self-rewarding prompt refinement. The overall framework is shown in Fig.~\ref{fig:overview}. 
We formulate prompt optimization as a closed-loop process that connects the user prompt, prompt optimizer, T2I generator, and generated image. 
Based on structured visual feedback, we introduce a two-stage training procedure consisting of supervised initialization and self-rewarding prompt optimization.

\subsection{Problem Formulation}
Given a user prompt $x$, prompt optimization aims to generate an improved prompt $\hat{x}$ that preserves the user intent while better adapting to a target T2I generator. 
Let $\pi_{\theta}$ denote the prompt optimization policy and $G$ denote the frozen T2I generator. 
The generation process is formulated as 
\begin{equation}
\hat{x} \sim \pi_{\theta}(\cdot \mid x), 
\quad
I = G(\hat{x}),
\end{equation}
where $I$ is the image generated from the optimized prompt.

To incorporate visual evidence into prompt optimization, we define a visual feedback function $F$ that produces both textual diagnosis and numerical scores:
\begin{equation}
\label{visualfeedback}
\begin{aligned}
\mathcal{F}
&= F(x,\hat{x},I) \\
&=
\left\{
(f_{\mathrm{sem}}, r_{\mathrm{sem}}),
(f_{\mathrm{aes}}, r_{\mathrm{aes}}),
(f_{\mathrm{pref}}, r_{\mathrm{pref}})
\right\},
\end{aligned}
\end{equation}
where $f_{\mathrm{sem}}$, $f_{\mathrm{aes}}$, and $f_{\mathrm{pref}}$ denote textual diagnostic feedback, and $r_{\mathrm{sem}}$, $r_{\mathrm{aes}}$, and $r_{\mathrm{pref}}$ denote the corresponding scores for semantic consistency, aesthetic quality, and preference alignment, respectively. The scores are collected into a reward vector
\begin{equation}
\mathbf{r}
=
\left[
r_{\mathrm{sem}},
r_{\mathrm{aes}},
r_{\mathrm{pref}}
\right],
\end{equation}
and further aggregated into a scalar reward by
\begin{equation}
R(x,\hat{x},I)=\Phi(\mathbf{r}).
\end{equation}

Thus, the objective of our framework is to learn a reusable prompt optimization policy by maximizing the expected visual-feedback reward:
\begin{equation}
\max_{\theta}
\mathbb{E}_{x \sim \mathcal{D},\, \hat{x} \sim \pi_{\theta}(\cdot \mid x)}
\left[
R(x,\hat{x},G(\hat{x}))
\right].
\end{equation}

\subsection{Visual Feedback Construction}
As defined in Equation~(\ref{visualfeedback}), visual feedback $\mathcal{F}$ contains both textual diagnosis and numerical scores along three dimensions: semantic consistency, aesthetic quality, and preference alignment. 
Semantic consistency measures whether the generated image preserves the original user intent, including objects, attributes, actions, counting, spatial relations, and interactions. 
Aesthetic quality evaluates visual appeal beyond semantic correctness, such as composition, lighting, color harmony, clarity, details, and artifacts. 
Preference alignment captures holistic user-oriented judgments, including naturalness, coherence, style appropriateness, attractiveness, and overall satisfaction. Details are provided in the Appendix~\ref{app:visual_feedbackscore}.

\subsection{Supervised Initialization for Prompt Rewriting and Visual Feedback Assessment}
\label{section:sft}
Before self-rewarding optimization, the VLM is initialized with two core abilities: prompt rewriting and structured visual feedback assessment. 
We train a single VLM under a multi-task supervised objective, enabling it to act as both the prompt optimizer and the feedback provider in the subsequent optimization stage.

\textbf{Training data construction.}
We construct the supervised dataset from four instruction types:
\begin{equation}
\mathcal{D}_{\mathrm{sup}}
=
\mathcal{D}_{\mathrm{rew}}
\cup
\mathcal{D}_{\mathrm{fb}}
\cup
\mathcal{D}_{\mathrm{score}}
\cup
\mathcal{D}_{\mathrm{fgr}} .
\end{equation}
The prompt rewriting set is defined as
\begin{equation}
\mathcal{D}_{\mathrm{rew}}
=
\{(x,\hat{x}^{*})\},
\end{equation}
where $x$ is the user prompt and $\hat{x}^{*}$ is the reference optimized prompt. 
The visual feedback assessment set is defined as
\begin{equation}
\mathcal{D}_{\mathrm{fb}}
=
\{(p,I_p,\mathcal{F}^{*}_p)\},
\
p\in\{x,\hat{x}^{*}\},
\
I_p=G(p),
\end{equation}
where $\mathcal{F}^{*}_p$ contains reference diagnostic feedback and dimension-specific scores. 
To calibrate numerical judgment, we further construct a score-only subset
\begin{equation}
\mathcal{D}_{\mathrm{score}}
=
\{(p,I_p,\mathbf{r}^{*}_p)\},
\end{equation}
where $\mathbf{r}^{*}_p=[r^{*}_{\mathrm{sem}},r^{*}_{\mathrm{aes}},r^{*}_{\mathrm{pref}}]$. 
Finally, the feedback guided rewriting set is defined as
\begin{equation}
\mathcal{D}_{\mathrm{fgr}}
=
\{(x,I_x,\mathcal{F}^{*}_x,\hat{x}^{*})\},
\end{equation}
which trains the model to revise under-specified prompts using explicit visual feedback. 
Detailed data construction procedures and instruction templates are provided in Appendix~\ref{app:data_processing}.

\textbf{Multi-task supervised fine-tuning.}
All training samples are converted into instruction-following formats and can be viewed in the Appendix~\ref{app:data_prompt_templates}. 
Let 
$\mathcal{T}=\{\mathrm{rew},\mathrm{fb},\mathrm{score},\mathrm{fgr}\}$ denote the set of supervised tasks, corresponding to prompt rewriting, visual feedback assessment, score-only calibration, and feedback-guided prompt rewriting, respectively. For each task $t \in \mathcal{T}$, we denote the instruction input as $u$ and the target response as $y$. The model is optimized with a response-only autoregressive objective:
\begin{equation}
\label{eq:sft_loss}
\mathcal{L}_{\mathrm{SFT}}
=
-
\sum_{t \in \mathcal{T}}
\lambda_t
\mathbb{E}_{(u,y)\sim \mathcal{D}_t}
\left[
\sum_{j=1}^{|y|}
\log \pi_{\theta}(y_j \mid u, y_{<j})
\right],
\end{equation}
where $\lambda_t$ controls the contribution of task $t$, and $\mathcal{D}_t$ denotes its corresponding training set. 

Task-specific instructions are used to specify the expected output format, including optimized prompts, diagnostic feedback with scores, score-only responses, and feedback-guided revised prompts. 
This multi-task initialization enables the same VLM to perform both prompt rewriting and visual feedback assessment, providing the starting policy for self-rewarding optimization.

\subsection{Self-Rewarding Prompt Optimization}
\label{section:self-reward}
After supervised initialization, the VLM is further optimized through self-rewarding training. 
In each iteration, the model acts as a prompt policy to generate candidate prompts and as a visual judge to provide reward signals. 
Given a user prompt $x$, we sample $K$ candidate prompts from the old policy:
\begin{equation}
\hat{x}_i \sim \pi_{\theta_{\mathrm{old}}}(\cdot \mid x),
\quad i=1,\ldots,K,
\end{equation}
and generate the corresponding images with the frozen T2I generator:
\begin{equation}
I_i = G(\hat{x}_i).
\end{equation}
A frozen copy of the initialized VLM is then used in feedback assessment mode to produce dimension-specific scores:
\begin{equation}
\mathcal{F}_i = F_{\theta_{\mathrm{old}}}(x,\hat{x}_i,I_i),
\quad
\mathbf{r}_i =
[r_{\mathrm{sem}}^i,r_{\mathrm{aes}}^i,r_{\mathrm{pref}}^i].
\end{equation}

\textbf{Hybrid reward aggregation.}
The multi-dimensional feedback scores are aggregated into a scalar reward for policy optimization. Since semantic consistency, aesthetic quality, and preference alignment capture complementary aspects of generation quality, optimizing a single dimension may lead to biased prompt improvement. We therefore adopt a hybrid reward aggregation strategy that combines the ideal-point distance with the Chebyshev distance.

The feedback scores are first normalized into $[0,1]$, yielding
$\bar{\mathbf{r}}_i=
[\bar r_{\mathrm{sem}}^i,\bar r_{\mathrm{aes}}^i,\bar r_{\mathrm{pref}}^i]$.
We define the ideal feedback vector as $\mathbf{r}^{*}=[1,1,1]$ and compute the weighted deviation:
\begin{equation}
\mathbf{d}_i
=
\mathbf{w}
\odot
(\mathbf{r}^{*}-\bar{\mathbf{r}}_i),
\end{equation}
where $\mathbf{w}=[w_{\mathrm{sem}},w_{\mathrm{aes}},w_{\mathrm{pref}}]$ controls the importance of different feedback dimensions. 
The scalar reward combines an ideal-point objective and a Chebyshev objective:
\begin{equation}
\begin{gathered}
R^{\mathrm{ideal}}_i
=
1-
\frac{\|\mathbf{d}_i\|_2}
{\|\mathbf{w}\odot\mathbf{r}^{*}\|_2}, \\[2mm]
R^{\mathrm{cheb}}_i
=
1-
\frac{\|\mathbf{d}_i\|_{\infty}}
{\|\mathbf{w}\odot\mathbf{r}^{*}\|_{\infty}}, \\[2mm]
R_i
=
\alpha R^{\mathrm{ideal}}_i
+
(1-\alpha)R^{\mathrm{cheb}}_i.
\end{gathered}
\end{equation}
Here, $R^{\mathrm{ideal}}_i$ encourages overall improvement toward the ideal feedback vector, while $R^{\mathrm{cheb}}_i$ penalizes the weakest dimension. 
The coefficient $\alpha\in[0,1]$ controls the trade-off between overall quality and worst-dimension penalty.

\textbf{Group-wise advantage estimation.}
To reduce absolute score bias, we compute relative advantages within each candidate group:
\begin{equation}
A_i
=
\frac{
R_i - \mathrm{mean}(\{R_j\}_{j=1}^{K})
}{
\mathrm{std}(\{R_j\}_{j=1}^{K})+\epsilon
}.
\end{equation}
This group-wise normalization encourages the policy to prefer prompts that achieve stronger visual feedback than other candidates sampled for the same input.

Since the reward is assigned to the whole optimized prompt, we adopt GSPO~\cite{zheng2025group}, which performs sequence-level rewarding, clipping, and policy optimization. 
For each candidate prompt $\hat{x}_i=(y_1,\ldots,y_T)$, the sequence-level importance ratio is defined as
\begin{equation}
\rho_i(\theta)
=
\exp
\left(
\frac{1}{T}
\sum_{t=1}^{T}
\log
\frac{
\pi_{\theta}(y_t \mid x,y_{<t})
}{
\pi_{\theta_{\mathrm{old}}}(y_t \mid x,y_{<t})
}
\right).
\end{equation}
We clip the ratio by
\begin{equation}
\tilde{\rho}_i(\theta)
=
\mathrm{clip}
\left(
\rho_i(\theta),
1-\epsilon_c,
1+\epsilon_c
\right),
\end{equation}
where $\epsilon_c$ is the sequence-level clipping threshold. The self-rewarding loss is
\begin{equation}
\begin{aligned}
\mathcal{L}_{\mathrm{SR}}
=
-&
\mathbb{E}_{x}
\left[
\frac{1}{K}
\sum_{i=1}^{K}
\min
\left(
\rho_i(\theta) A_i,
\tilde{\rho}_i(\theta) A_i
\right)
\right] \\
&+
\beta
\mathcal{D}_{\mathrm{KL}}
\left(
\pi_{\theta}
\|
\pi_{\mathrm{ref}}
\right),
\end{aligned}
\end{equation}
where $\pi_{\mathrm{ref}}$ denotes the supervised initialized reference policy and $\beta$ controls the KL regularization strength.

This self-rewarding stage reinforces prompts that receive stronger multi-dimensional visual feedback, enabling the policy to progressively improve prompt optimization without relying on a fixed external reward model.

\section{Experiments}
\begin{table*}[t]
  \centering
  \caption{Holistic image-level evaluation on the BeautifulPrompt test set. PRISM-SFT denotes the supervised-initialized variant without self-rewarding optimization, and PRISM denotes the full model. Best and second-best results within each group are highlighted in bold and underlined, respectively.}
  \label{tab:beautifulprompt_results}
  \small
  \setlength{\tabcolsep}{5.5pt}
  \renewcommand{\arraystretch}{1.12}
  \begin{tabular}{@{}lccccc@{}}
    \toprule
    Method & CLIPScore $\uparrow$ & Aesthetic $\uparrow$ & PickScore $\uparrow$ & HPSv2 $\uparrow$ & ImageReward $\uparrow$ \\
    \midrule
    
    Original Prompt
    & 0.282 & 6.106 & 21.219 & 0.276 & 0.989 \\
    \midrule

    \rowcolor{gray!12}
    \multicolumn{6}{@{}l}{\textit{Text-only prompt optimization}} \\
    BestPrompt
    & 0.268 & 6.182 & 20.143 & 0.260 & 0.914 \\
    BeautifulPrompt
    & 0.272 & \uline{6.291} & 21.088 & 0.277 & 1.050 \\
    Promptist
    & \uline{0.283} & 6.263 & 21.506 & 0.281 & 1.019 \\
    NeuroPrompts
    & 0.277 & 6.219 & \uline{22.195} & 0.285 & 1.112 \\
    Qwen3-VL-8B
    & 0.245 & 6.136 & 21.128 & 0.278 & 1.195 \\
    PRISM-SFT
    & 0.280 & 6.198 & 22.149 & \uline{0.297} & \uline{1.201}\\
    PRISM
    & \textbf{0.291} & \textbf{6.307} & \textbf{22.457} & \textbf{0.302} & \textbf{1.218} \\
    \midrule

    \rowcolor{gray!12}
    \multicolumn{6}{@{}l}{\textit{Text-image feedback-based prompt optimization}} \\
    VisualPrompter
    & \textbf{0.304} & 6.131 & 21.208 & 0.288 & 0.993 \\
    TIR
    & 0.285 & 6.176 & 21.657 & 0.280 & 0.972 \\
    OPT2I
    & 0.291 & 6.150 & 21.501 & 0.272 & 1.006 \\
    GPT-4o
    & 0.275 & \uline{6.255} & 21.955 & 0.283 & 1.198 \\
    Qwen3-VL-235B-A22B
    & 0.269 & 6.172 & \uline{22.381} & 0.289 & \uline{1.269} \\
    Qwen3-VL-8B
    & 0.261 & 6.129 & 21.628 & 0.288 & 1.176 \\
    PRISM-SFT
    & 0.278 & 6.223 & 22.254 & \uline{0.303} & 1.252 \\
    PRISM
    & \uline{0.299} & \textbf{6.318} & \textbf{22.526} & \textbf{0.305} & \textbf{1.285} \\
    \bottomrule
  \end{tabular}
\end{table*}
We evaluate PRISM on T2I prompt optimization. 
Our experiments examine whether the learned prompt policy improves image-text alignment, visual quality, and human preference alignment, and whether visual feedback and self-rewarding optimization provide consistent gains over supervised initialization.

\subsection{Experimental Settings}

We adopt Qwen3-VL-8B-Instruct~\cite{bai2025qwen3} as the backbone VLM.
Stable Diffusion 3.5 medium\footnote{\url{https://huggingface.co/stabilityai/stable-diffusion-3.5-medium}} is used as the primary target T2I generator with 28 denoising steps, while additional results on FLUX.2-klein\footnote{\url{https://huggingface.co/black-forest-labs/FLUX.2-klein-4B}} are reported in Appendix~\ref{app:flux_results} to evaluate cross-generator generalization.

\textbf{Supervised fine-tuning.}
We train the VLM with AdamW~\cite{loshchilov2017decoupled} using $\beta_1=0.9$, $\beta_2=0.95$, a batch size of 32, and weight decay of 0.1. The learning rate is initialized to $2\times10^{-5}$ and scheduled with linear warm-up followed by cosine decay. For the multi-task SFT objective, we set the task loss weights to 
$\lambda_{\mathrm{rew}}:\lambda_{\mathrm{fb}}:\lambda_{\mathrm{score}}:\lambda_{\mathrm{fgr}}=1:1:1:2$ emphasizing feedback-guided prompt rewriting.

\textbf{Self-rewarding optimization.}
For each input prompt, the policy samples $K=8$ candidate prompts with temperature $T=0.9$ and top-$p=0.9$. 
We set the sequence-level clipping threshold to $\epsilon_c=0.2$ and the KL coefficient to $\beta=0.3$, using the SFT model as the reference policy $\pi_{\mathrm{ref}}$. 
The scalar reward follows the hybrid aggregation strategy in Section~\ref{section:self-reward}, with $\mathbf{w}=[0.5,0.25,0.25]$ to emphasize semantic consistency. Each visual assessment is repeated three times and averaged for stability. 
We perform two self-rewarding iterations. 

During training and evaluation, all T2I generators are frozen. 
All methods are evaluated under the same generation settings for fair comparison. 
All experiments are conducted on NVIDIA A800 80G GPUs.

\subsection{Benchmarks and Evaluation Metrics}
We evaluate PRISM on two types of benchmarks. 
For general prompt optimization, we use the 2k BeautifulPrompt test prompts~\cite{cao2023beautifulprompt} and report CLIPScore~\cite{hessel2021clipscore}, Aesthetic Score~\cite{podell2024sdxl}, ImageReward~\cite{xu2023imagereward}, PickScore~\cite{kirstain2023pick}, and HPSv2~\cite{wu2023human} to measure image-text alignment, visual quality, and human preference. 
For fine-grained semantic alignment, we evaluate on T2I-CompBench~\cite{huang2023t2i} and TIFA~\cite{hu2023tifa}. 
T2I-CompBench assesses compositional generation over attributes, relations, and complex prompts, while TIFA measures prompt-image faithfulness via automatically generated visual questions. 
For all benchmarks, metrics are computed with respect to the original user prompt rather than the optimized prompt, ensuring that the evaluation reflects preservation and enhancement of the original user intent.

\subsection{Comparative Results}
\begin{table*}[t]
  \centering
  \caption{Fine-grained semantic alignment evaluation on TIFA and T2I-CompBench. Best and second-best results are highlighted in bold and underlined, respectively.}
  \label{tab:fine_grained_results}
  \small
  \setlength{\tabcolsep}{4.0pt}
  \renewcommand{\arraystretch}{1.12}
  \begin{tabular}{@{}lcccccccc@{}}
    \toprule
    \multirow{2}{*}{Method}
    & \multirow{2}{*}{TIFA $\uparrow$}
    & \multicolumn{7}{c}{T2I-CompBench $\uparrow$} \\
    \cmidrule(lr){3-9}
    & & Color & Shape & Texture & Spatial & NonSpatial & Complex & Avg. \\
    \midrule
    Original Prompt
    & 76.1 & 0.736 & 0.510 & 0.627 & 0.259 & 0.265 & 0.298 & 0.449 \\
    
    \midrule
    VisualPrompter
    & 87.8 & 0.791 & \textbf{0.605} & 0.692 & 0.403 & 0.315 & \uline{0.402} & \uline{0.535} \\
    TIR
    & 82.4 & 0.820 & 0.559 & 0.715 & 0.355 & 0.309 & 0.384 & 0.524 \\
    OPT2I
    & 85.1 & 0.812 & 0.592 & \uline{0.724} & 0.386 & 0.314 & 0.372 & 0.533 \\
    GPT-4o
    & \uline{90.3} & 0.835 & 0.548 & 0.713 & \uline{0.415} & 0.302 & 0.377 & 0.532 \\
    Qwen3-VL-235B-A22B
    & 89.6 & 0.808 & 0.577 & 0.708 & 0.392 & \textbf{0.318} & 0.395 & 0.533 \\
    Qwen3-VL-8B
    & 87.5 & 0.754 & 0.580 & 0.682 & 0.371 & 0.313 & 0.364 & 0.511 \\
    PRISM-SFT
    & 88.9 & 0.793 & 0.585 & 0.704 & 0.385 & 0.311 & 0.396 & 0.529 \\
    PRISM
    & \textbf{91.4} & \textbf{0.843} & \uline{0.603} & \textbf{0.732} & \textbf{0.429} & \uline{0.316} & \textbf{0.405} & \textbf{0.555} \\
    \bottomrule
  \end{tabular}

  \vspace{1mm}
  \begin{minipage}{0.96\textwidth}
  \footnotesize
  \end{minipage}
\end{table*}
\textbf{Holistic image-level evaluation.}
Table~\ref{tab:beautifulprompt_results} reports the main results on the BeautifulPrompt test set. 
In the text-only setting, PRISM achieves the best results across all five metrics, showing consistent gains over both original prompts and prior prompt optimization methods. 
In the text-image feedback setting, PRISM obtains the best Aesthetic, PickScore, HPSv2, and ImageReward scores, while remaining competitive on CLIPScore. 
These results indicate that visual-feedback-driven optimization improves prompt quality in a balanced manner, enhancing not only image-text alignment but also aesthetic quality and human preference alignment.

\textbf{Fine-grained semantic alignment.}
Table~\ref{tab:fine_grained_results} reports the results on TIFA and T2I-CompBench. 
PRISM achieves the best overall performance, improving TIFA from 76.1 to 91.4 and the T2I-CompBench average from 0.449 to 0.555 compared with the original prompts. 
It also surpasses the strongest baselines on both TIFA and the average compositional score, and obtains the best results on color, texture, spatial relation, and complex composition. 
Moreover, the gains over PRISM-SFT demonstrate that the self-rewarding stage further strengthens fine-grained semantic alignment by leveraging structured visual feedback, particularly for attribute binding, spatial reasoning, and complex compositions.

\subsection{Ablation Study}
\begin{table}[htbp]
  \centering
  \caption{Ablation on feedback sources. PS denotes PickScore. FB denotes feedback source. 1R/2R denote one/two test-time refinement rounds. $C_{\mathrm{T2I}}$ and $C_{\mathrm{VLM}}$ denote the numbers of T2I and VLM calls per input prompt.}
  \label{tab:feedback_source_ablation}
  \small
  \setlength{\tabcolsep}{2.1pt}
  \renewcommand{\arraystretch}{1.08}
  \begin{tabular}{@{}lcccccc@{}}
    \toprule
    Method & FB & Train & PS $\uparrow$ & TIFA $\uparrow$ & $C_{\mathrm{T2I}}\downarrow$ & $C_{\mathrm{VLM}}\downarrow$ \\
    \midrule
    TF-Ext (1R) & Ext  & No          & 20.984 & 86.3 & 2 & 1 \\
    TF-Ext (2R) & Ext  & No          & 21.343 & 88.2 & 3 & 2 \\
    Self-FB     & Self & SFT         & 22.254 & 88.9 & 2 & 1 \\
    Ext-VLM     & Ext  & RL  & \uline{22.411} & \uline{90.0} & 2 & 1 \\
    PRISM        & Self & RL  & \textbf{22.526} & \textbf{91.4} & 2 & 1 \\
    \bottomrule
  \end{tabular}
\end{table}
\textbf{Effect of feedback source.}
Table~\ref{tab:feedback_source_ablation} shows that learning with feedback is more effective than training-free refinement. 
Under the same inference cost as TF-Ext (1R), Self-FB improves PS from 20.984 to 22.254 and TIFA from 86.3 to 88.9. 
With self-rewarding optimization, PRISM further achieves the best results on both PS and TIFA, outperforming Ext-VLM under the same cost. 
This suggests that integrating self-feedback into the trainable policy provides more effective and consistent reward signals.
\begin{figure}[htbp]
  \includegraphics[width=\columnwidth]{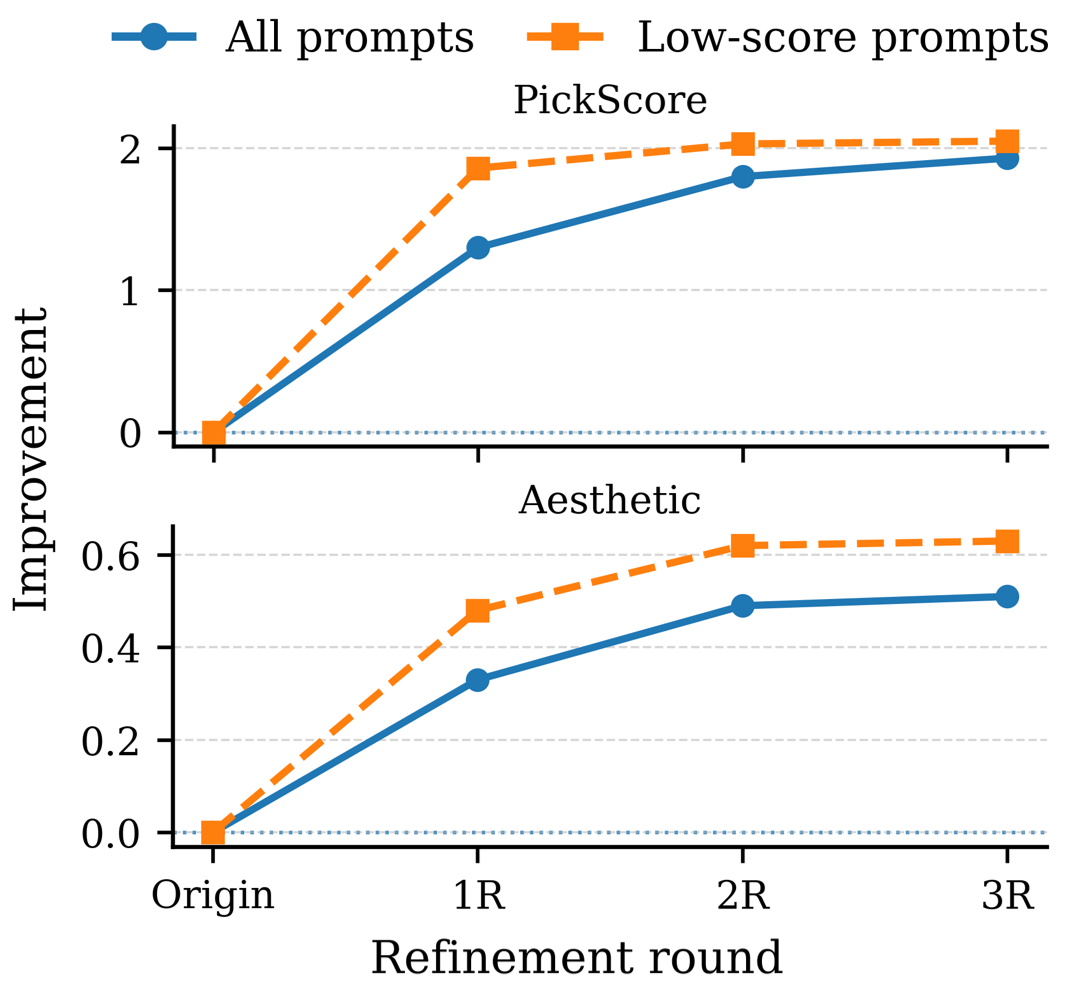}
  \caption{Iterative feedback refinement results across multiple refinement rounds.}
  \label{figs:iterative_refinement}
\end{figure}

\begin{figure*}[htbp]
  \includegraphics[width=1.0\linewidth]{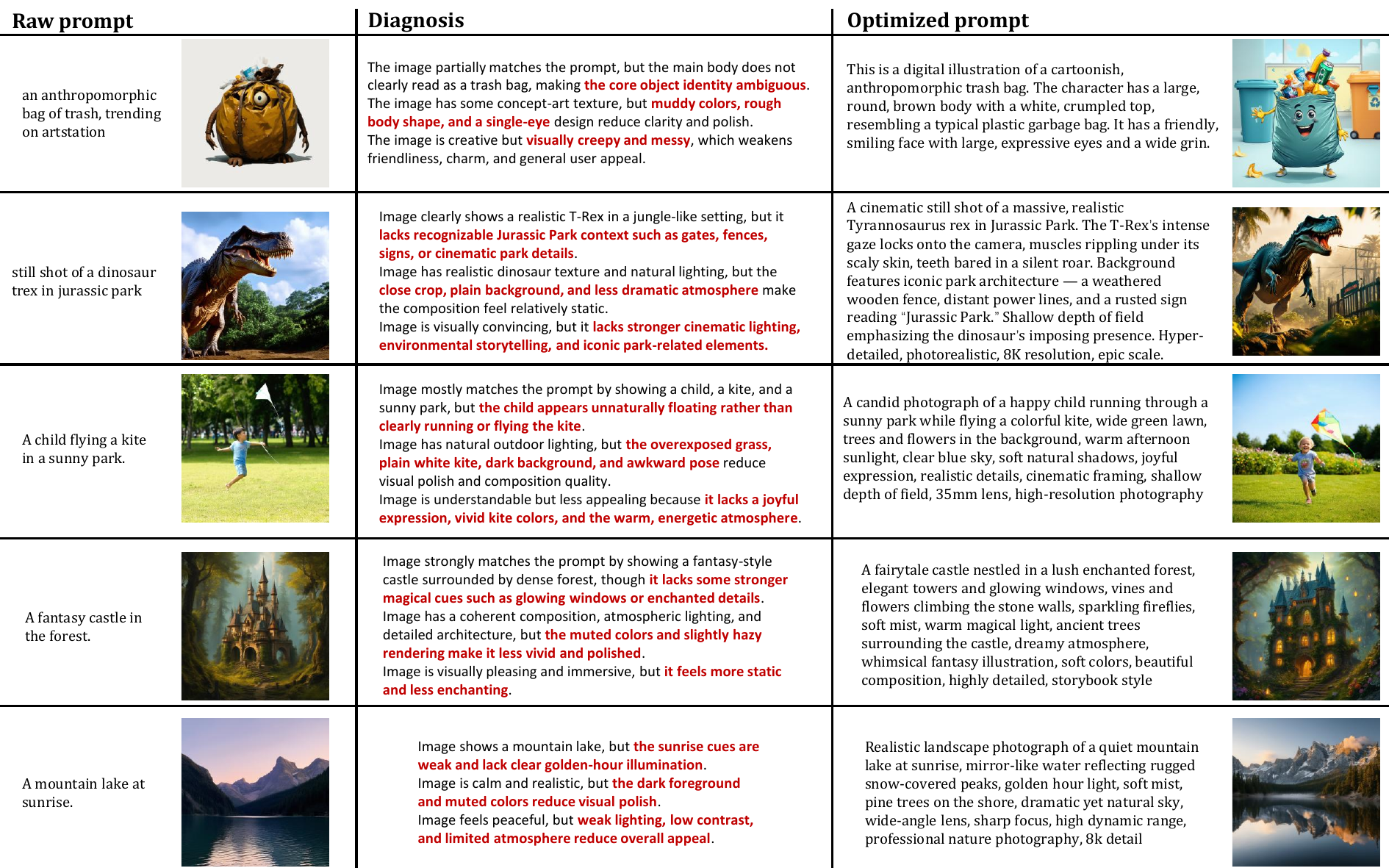}
  \caption{Qualitative examples of PRISM. 
  Each example shows the original prompt, initial generated image, structured visual feedback, optimized prompt, and final generated image, illustrating how diagnosed visual issues guide targeted prompt refinement.}
  \label{fig:qualitative_examples}
\end{figure*}
\textbf{Iterative feedback refinement.}
We further evaluate whether the model can benefit from repeated visual feedback. 
Starting from the original prompt, we perform multiple refinement rounds by feeding the generated image and its visual feedback back into the model. 
As shown in Fig.~\ref{figs:iterative_refinement}, the second refinement round brings additional improvements, especially for samples with low first-pass scores. 
This suggests that PRISM can act as a feedback-aware prompt optimizer rather than a one-shot prompt enhancer.

\textbf{Effect of reward aggregation.}
Table~\ref{tab:reward_ablation} compares different reward aggregation strategies. 
The hybrid reward achieves the best performance across all metrics.
Compared with the weighted-sum reward, the hybrid strategy further improves TIFA by 3.3 points while maintaining slightly better aesthetic and preference scores. 
Although the ideal-point and Chebyshev rewards improve semantic faithfulness over the weighted-sum baseline, they show weaker aesthetic or preference performance when used alone. 
These results indicate that combining overall closeness to the ideal feedback profile with worst-dimension penalty leads to more balanced prompt optimization.
\begin{table}[htbp]
  \centering
  \caption{Ablation on reward aggregation strategies.}
  \label{tab:reward_ablation}
  \small
  \setlength{\tabcolsep}{4.0pt}
  \renewcommand{\arraystretch}{1.12}
  \begin{tabular}{@{}lccc@{}}
    \toprule
    Reward & Aesthetic $\uparrow$ & PickScore $\uparrow$ & TIFA $\uparrow$ \\
    \midrule
    Weighted Sum
    & \uline{6.312} & \uline{22.512} & 88.1 \\
    Ideal-point
    & 6.281 & 22.495 & \uline{89.8} \\
    Chebyshev
    & 6.275 & 22.403 & 89.2 \\
    Hybrid
    & \textbf{6.318} & \textbf{22.526} & \textbf{91.4} \\
    \bottomrule
  \end{tabular}

  \vspace{1mm}
  \begin{minipage}{0.95\linewidth}
  \footnotesize
  \end{minipage}
\end{table}

\subsection{Qualitative Analysis}
Fig.~\ref{fig:qualitative_examples} presents representative examples of the visual-feedback-guided prompt refinement process in PRISM. 
The model identifies concrete generation issues, such as missing objects, weak spatial relations, and style mismatch, and revises the prompt accordingly. 
These examples demonstrate that structured visual feedback provides interpretable correction cues for targeted prompt refinement.

\section{Conclusion}
In this paper, we propose PRISM, a Prompt Refinement framework via Image-grounded Self-rewarding Mechanism for text-to-image prompt optimization. 
PRISM uses structured visual diagnosis to guide prompt rewriting and reward computation across semantic consistency, aesthetic quality, and preference alignment. 
Through multi-task supervised initialization and self-rewarding policy optimization, the VLM learns to jointly perform prompt rewriting and visual feedback assessment. 
Extensive experiments demonstrate that PRISM improves holistic image quality and fine-grained semantic alignment, while providing interpretable feedback for targeted prompt refinement.

\section*{Limitations}

Although PRISM demonstrates strong effectiveness, several aspects remain open for future work. First, the current feedback design focuses on semantic consistency, aesthetic quality, and preference alignment; extending it to more fine-grained or domain-specific criteria could further improve controllability. Second, iterative feedback refinement is evaluated with a limited number of rounds to maintain inference efficiency, while adaptive stopping strategies may provide a better trade-off between quality and cost. 
\bibliography{custom}

\clearpage
\appendix

\section{Data Processing Pipeline}
\label{app:data_processing}

We construct four types of training data for PRISM:
(i) rewriting data $\mathcal D_{\mathrm{rew}}$,
(ii) visual feedback data $\mathcal D_{\mathrm{vf}}$,
and (iii) score-only calibration data $\mathcal D_{\mathrm{score}}$.
The rewriting data are sampled from BeautifulPrompt.
For visual feedback data, we first build a tractable candidate pool from open community T2I data sources, including DiffusionDB, JourneyDB, and Pick-a-Pic.
We then select high-quality image-text pairs from this candidate pool and construct degraded prompts through controlled simplification.
The resulting data support three objectives:
basic prompt rewriting $x\rightarrow\hat{x}$,
visual feedback estimation $F(x,\hat{x},I)$,
and feedback-conditioned prompt refinement $(x,I,\mathcal F)\rightarrow\hat{x}$.

\subsection{Unified Quality Score}

Let $a_i$, $p_i$, and $c_i$ denote the Aesthetic score, PickScore, and CLIPScore of the $i$-th sample, respectively.
For a candidate pool $\mathcal U$, each metric is normalized by pool-wise min--max scaling:
\begin{equation}
\tilde m_i
=
\frac{m_i-\min_{j\in\mathcal U} m_j}
{\max_{j\in\mathcal U} m_j-\min_{j\in\mathcal U} m_j+\epsilon},
\label{eq:minmax}
\end{equation}
where $m\in\{a,p,c\}$ and $\epsilon=10^{-8}$.
We then define a unified quality score:
\begin{equation}
S_i
=
0.25\,\tilde a_i
+
0.25\,\tilde p_i
+
0.50\,\tilde c_i.
\label{eq:quality_score}
\end{equation}
The larger CLIPScore weight emphasizes image-text semantic alignment.
We use Equation(~\eqref{eq:quality_score}) for ranking samples in both Type-I and Type-II data construction.

\subsection{BeautifulPrompt Rewriting Data}

BeautifulPrompt contains approximately $143$k prompt pairs.
Since its training metadata already provides Aesthetic, PickScore, and CLIPScore annotations, we do not perform additional external rescoring.
We rank the BeautifulPrompt training pool $\mathcal U_{\mathrm{BP}}$ using Eqs.~\eqref{eq:minmax}--\eqref{eq:quality_score}, and retain the top $N_1=10{,}000$ pairs, corresponding to about $7.0\%$ of the original training pool.

The resulting Type-I dataset is $\mathcal D_{\mathrm{rew}}=\{(x_i,\hat{x}_i)\}_{i=1}^{N_1}$, where $x_i$ is a low-quality prompt, and $\hat{x}_i$ is its high-quality rewritten version.
This dataset is used only for supervised initialization of the prompt optimizer $\pi_\theta$, i.e., $x\rightarrow\hat{x}$.

\subsection{Community Data and Controlled Prompt Simplification}

\subsubsection{Candidate pool construction}

We construct feedback-oriented data from publicly available and license-permissive AI-generated image resources.
Rather than processing full-scale community datasets, we first sample a tractable candidate pool
$\mathcal U_{\mathrm{T2I}}$ with $C$ image-text pairs.
In our implementation, we set $C=50{,}000$ and perform all metric computation and filtering only within this sampled pool.
The pool can be instantiated from open community T2I resources, including DiffusionDB, JourneyDB, and Pick-a-Pic, depending on data availability and experimental needs.

Each candidate sample is represented as $(\hat{x}_i,I_i^{+})$, where $\hat{x}_i$ is a community-written prompt and $I_i^{+}$ is its associated generated image.
Since community prompts often contain rich visual details and generation-oriented expressions, we treat $\hat{x}_i$ as a model-preferred prompt.
For each candidate pair, we compute three quality indicators:
$a_i$ for aesthetic quality, $p_i$ for preference alignment, and $c_i$ for image-text alignment.
After min--max normalization within $\mathcal U_{\mathrm{T2I}}$, we compute the unified quality score in Equation(~\eqref{eq:quality_score}) and retain the top $N_2=5{,}000$ high-quality pairs.

The retained high-quality pool is denoted as
$\mathcal D_{\mathrm{high}}=\{(\hat{x}_i,I_i^{+})\}_{i=1}^{N_2}$.
This pool serves as the source for constructing paired prompts $(x_i,\hat{x}_i)$ and their feedback-oriented supervision.

\subsubsection{Controlled prompt simplification}

For each high-quality prompt $\hat{x}_i$, we apply a stochastic simplification operator $\mathcal T$ to construct an under-specified counterpart
$x_i=\mathcal T(\hat{x}_i)$.
The goal is not to introduce semantic errors, but to preserve the core intent while removing details that are commonly missing in user-provided prompts.
This yields paired prompts $(x_i,\hat{x}_i)$ with explicit optimization gaps.

The simplification process follows four primary strategies:
\begin{itemize}
    \item \textbf{Summarization and abstraction ($30\%$).}
    We compress a long prompt into a short sentence or tag chain, with target length ratio $r\sim\mathcal U(0.25,0.50)$.
    Fine-grained nouns may be replaced by more generic hypernyms, while the main entity and action are retained.
    For example, \texttt{a Victorian gothic cathedral, dramatic backlight, intricate stone texture}
    is simplified to \texttt{an old cathedral with dramatic light}.

    \item \textbf{Removal of style and quality modifiers ($25\%$).}
    We remove common style, quality, and rendering modifiers, such as
    \texttt{masterpiece, best quality, ultra-detailed, 8k, RAW photo, cinematic lighting, octane render, artstation, unreal engine, sharp focus}.
    Specifically, we randomly delete $60\%\!\sim\!100\%$ of matched modifiers.
    For example, \texttt{masterpiece, ultra detailed, 8k, cinematic lighting, 1 girl}
    is reduced to \texttt{1 girl}.

    \item \textbf{Weakening attributes, relations, scenes, and composition ($30\%$).}
    We decompose $\hat{x}_i$ into atomic semantic units and remove $1\!\sim\!3$ non-core atoms, preferably attributes, relations, scene details, and composition descriptions.
    This includes removing color, material, texture, clothing, spatial relations, exact counts, camera views, and framing details.
    For example, \texttt{three red apples on the left side of a wooden table, close-up shot}
    is simplified to \texttt{apples on a table}.

    \item \textbf{Omission of secondary entities ($15\%$).}
    When a prompt contains multiple entities, background objects, or auxiliary events, we keep the main entity and at most one supporting entity, while removing peripheral clauses and minor interactions.
    For example, \texttt{a boy reading near a window, with a cat on the sofa and plants in the background}
    is simplified to \texttt{a boy reading near a window}.
\end{itemize}

To improve degradation diversity, we additionally apply one secondary micro-edit with probability $0.35$.
During simplification, we enforce three constraints: at least one main entity is preserved, the token retention ratio satisfies $r\in[0.25,0.70]$, and the degraded prompt is non-empty.

For each paired prompt sample $(x_i,\hat{x}_i)$, we associate images with prompts $p\in\{x_i,\hat{x}_i\}$.
Specifically, $I_i$ is generated by the frozen target T2I generator $G$, i.e., $I_i=G(x_i)$.
For the optimized prompt $\hat{x}_i$, we reuse the original community image $I_i^{+}$. This produces intermediate prompt-image samples of the form $\{x_i, I_i, \hat{x}_i, I_i^{+}\}$.
These samples are then organized into the feedback assessment, score-only calibration, and feedback-guided rewriting datasets described below.

\subsection{Visual Feedback Annotation Pipeline}
\label{app:visual_feedbackscore}

\paragraph{Semantic feedback.}
Semantic feedback evaluates whether the generated image $I_p$ preserves the semantic intent specified by the reference prompt $p$.
Similar to Davidsonian Scene Graph(DSG)-based semantic prompt analysis, we decompose the reference prompt into atomic semantic units and compare them with the image through question answering.
When $p=x_i$, we use $\hat{x}_i$ as the reference prompt, so that the feedback can reveal semantic and compositional details missing from the under-specified prompt.
When $p=\hat{x}_i$, the optimized prompt itself is used as the reference.

Concretely, we first use Qwen3 to parse the reference prompt into DSG-style atomic units and convert them into yes/no questions.
The questions cover entity, attribute, relation, count, text, and spatial concepts.
Given $I_p$, Qwen3-VL answers each question to obtain the predicted label $\hat y_q\in\{0,1\}$, where $y_q\in\{0,1\}$ denotes the expected answer.

We compute the semantic score by dependency-aware question-answering accuracy:
\begin{equation}
\bar r_{p,\mathrm{sem}}
=
\frac{1}{|Q|}
\sum_{q\in Q}
\mathbbm{1}[\hat y_q = y_q],
\quad
\bar r_{p,\mathrm{sem}}\in[0,1].
\label{eq:sem_score}
\end{equation}
The atomic comparison results $\{(q,y_q,\hat y_q)\}_{q\in Q}$ are then summarized by Qwen3 into concise textual diagnostic feedback $f^{*}_{p,\mathrm{sem}}$.
The final semantic score $r^{*}_{p,\mathrm{sem}}\in\{1,\dots,10\}$ is obtained by discretizing $\bar r_{p,\mathrm{sem}}$ in Appendix~\ref{app:discretization}.

\paragraph{Aesthetic feedback.}
Aesthetic feedback evaluates the visual quality of the generated image $I_p$.
We use Qwen3-VL to produce textual diagnostic feedback $f^{*}_{p,\mathrm{aes}}$ according to a fixed rubric:
\emph{composition, lighting, color harmony, image clarity, visual details, artifacts}.
Qwen3-VL is only used for diagnosis generation and does not provide the final supervised score.

For numerical assessment, we compute an aesthetic metric score $a_p$ for $I_p$ and normalize it:
$\bar r_{p,\mathrm{aes}}=\mathrm{mm}(a_p;\mathcal D_{\mathrm{high}}),$
where $\mathrm{mm}(\cdot)$ denotes the min--max normalization in Equation(~\eqref{eq:minmax}).
The normalized score $\bar r_{p,\mathrm{aes}}\in[0,1]$ is used only as an intermediate continuous score.
The final aesthetic score $r^{*}_{p,\mathrm{aes}}\in\{1,\dots,10\}$ is obtained by quantile-based discretization in Appendix.~\ref{app:discretization}.

\paragraph{Preference feedback.}
Preference feedback captures overall user satisfaction and preference alignment.
The textual feedback $f^{*}_{p,\mathrm{pref}}$ is generated from five aspects:
\emph{naturalness, coherence, style appropriateness, attractiveness, overall satisfaction}.
The numerical preference score is computed by averaging normalized scores from multiple preference scorers:
\begin{equation}
\bar r_{p,\mathrm{pref}}
=
\frac{
\tilde r^{\mathrm{IR}}_p+\tilde r^{\mathrm{PS}}_p+\tilde r^{\mathrm{HPS}}_p
}{3},
\quad
\bar r_{p,\mathrm{pref}}\in[0,1].
\label{eq:pref_score}
\end{equation}
where $\tilde r^{\mathrm{IR}}_p$, $\tilde r^{\mathrm{PS}}_p$, and $\tilde r^{\mathrm{HPS}}_p$ denote normalized ImageReward, PickScore, and HPSv2 scores, respectively. The final preference score $r^{*}_{p,\mathrm{pref}}\in\{1,\dots,10\}$ is obtained by quantile-based discretization in Appendix.~\ref{app:discretization}.
This design decouples interpretable textual critique from reproducible numerical supervision.

\begin{table*}[t]
\centering
\small
\setlength{\tabcolsep}{5pt}
\begin{tabular}{lccc}
\toprule
Dataset & Source & Sample Format & Size \\
\midrule
$\mathcal D_{\mathrm{rew}}$ &
BeautifulPrompt &
$(x,\hat{x}^{*})$ &
$10{,}000$ \\
$\mathcal D_{\mathrm{fb}}$ &
Sampled community T2I pool &
$(p,I_p,\mathcal F^{*}_p)$ &
$10{,}000$ \\
$\mathcal D_{\mathrm{score}}$ &
Generated or reused images &
$(p,I_p,\mathbf r^{*}_p)$ &
$10,000$ \\
$\mathcal D_{\mathrm{fgr}}$ &
Sampled community T2I pool $+$ target T2I &
$(x,I_x,\mathcal F^{*}_x,\hat{x})$ &
$5{,}000$ \\
\bottomrule
\end{tabular}
\caption{
Summary of the training datasets.
Here $p\in\{x,\hat{x}\}$, $\mathcal F^{*}_p$ denotes reference visual feedback, and
$\mathbf r^{*}_p=[r^{*}_{p,\mathrm{sem}},r^{*}_{p,\mathrm{aes}},r^{*}_{p,\mathrm{pref}}]$.
The size $M$ depends on the number of samples included for score-only calibration.
}
\label{tab:data_counts}
\end{table*}
\subsection{Discretization}
\label{app:discretization}

The feedback annotation pipeline first produces continuous intermediate scores
$\bar{\mathbf r}_p=[\bar r_{p,\mathrm{sem}},\bar r_{p,\mathrm{aes}},\bar r_{p,\mathrm{pref}}]\in[0,1]^3$.
For supervision and reward computation, we discretize them into integer scores
$\mathbf r^{*}_p=[r^{*}_{p,\mathrm{sem}},r^{*}_{p,\mathrm{aes}},r^{*}_{p,\mathrm{pref}}]\in\{1,\dots,10\}^3$.
Thus, $\mathbf r^{*}_p$ denotes the final reference score vector used in
$\mathcal D_{\mathrm{fb}}$, $\mathcal D_{\mathrm{score}}$, and reward aggregation.

For semantic consistency, we use uniform bins over $[0,1]$:
\begin{equation}
r^{*}_{p,\mathrm{sem}}
=
\begin{cases}
10, & \bar r_{p,\mathrm{sem}}=1,\\[2mm]
\lfloor 10\bar r_{p,\mathrm{sem}} \rfloor + 1, & 0 \le \bar r_{p,\mathrm{sem}} < 1.
\end{cases}
\label{eq:sem_bin}
\end{equation}

For aesthetic quality and preference alignment, we use training-set quantile bins to reduce distribution imbalance.
For each $u\in\{\bar r_{p,\mathrm{aes}},\bar r_{p,\mathrm{pref}}\}$, we compute nine decile thresholds on the training split:
$e_k(u)=\operatorname{Quantile}(u_{\mathrm{train}},0.1k)$, $k=1,\dots,9$.
The discretized score is then defined as
\begin{equation}
r^{*}_{p,d}
=
1+\sum_{k=1}^{9}\mathbbm{1}[\bar r_{p,d} \ge e_k(d)],
\quad
d\in\{\mathrm{aes},\mathrm{pref}\}.
\label{eq:quantile_bin}
\end{equation}
All final scores are therefore integers in $\{1,\dots,10\}$, where larger values indicate better semantic consistency, aesthetic quality, or preference alignment.

\subsection{Supervision Formats}

The constructed samples are organized into four supervision formats, consistent with the main paper. Table~\ref{tab:data_counts} summarizes the training datasets.

\paragraph{Prompt rewriting data.}
The first type is prompt rewriting data:
\begin{equation}
\mathcal D_{\mathrm{rew}}
=
\{(x,\hat{x}^{*})\}.
\end{equation}
We use a curated $10$k-sample subset of BeautifulPrompt.
This dataset teaches the prompt optimizer $\pi_\theta$ to transform simple or under-specified user prompts into more detailed and model-preferred prompts.

\paragraph{Visual feedback assessment data.}
The second type is visual feedback assessment data:
\begin{equation}
\mathcal D_{\mathrm{fb}}
=
\{(p,I_p,\mathcal F^{*}_p)\}.
\end{equation}
Here, $p\in\{x_i,\hat{x}_i\}$, $I_p$ is the generated or reused image associated with prompt $p$, and $\mathcal F^{*}_p$ contains structured diagnostic feedback and scores over semantic consistency, aesthetic quality, and preference alignment.
This dataset initializes the model's ability to assess generated images and produce interpretable feedback.

\paragraph{Score-only calibration data.}
To stabilize numerical scoring, we additionally construct score-only calibration data:
\begin{equation}
\mathcal D_{\mathrm{score}}
=
\{(p,I_p,\mathbf r^{*}_p)\}.
\end{equation}
This subset removes long-form diagnostic text and retains only the score vector
$\mathbf r^{*}_p=[r^{*}_{p,\mathrm{sem}},r^{*}_{p,\mathrm{aes}},r^{*}_{p,\mathrm{pref}}]$.
It prevents score prediction from being overwhelmed by free-form diagnostic supervision.

\paragraph{Feedback-guided prompt rewriting data.}
The final instruction type is feedback-guided prompt rewriting data:
\begin{equation}
\mathcal D_{\mathrm{fgr}}
=
\{(x_i,I_i,\mathcal F^{*}_{x_i},\hat{x}_i)\}.
\end{equation}
Unlike $\mathcal D_{\mathrm{fb}}$, this task explicitly connects visual diagnosis with prompt revision.
Given an under-specified prompt $x_i$, its generated image $I_i=G(x_i)$, and its visual feedback $\mathcal F^{*}_{x_i}$, the model is trained to revise the prompt toward the optimized counterpart $\hat{x}_i$.
This encourages the model to use visual feedback as explicit correction cues rather than treating image assessment and prompt rewriting as isolated tasks.

\section{Training Data Prompt Templates}
\label{app:data_prompt_templates}

We provide the prompt templates used to construct the four types of training data. Each template is written in an instruction-style format, matching the actual input-output structure used during data generation.
\tcbset{
  promptbox/.style={
    width=\linewidth,
    colback=gray!5,
    colframe=gray!60,
    fonttitle=\bfseries,
    breakable
  }
}

\begin{tcolorbox}[
  title={Prompt Rewriting Template for $\mathcal D_{\mathrm{rew}}$},
  width=\linewidth,
  colback=gray!5,
  colframe=gray!60,
  fonttitle=\bfseries,
  breakable
]
Please rewrite the following T2I prompt into a higher-quality prompt.

The rewritten prompt should:
\begin{itemize}
  \item preserve the original user intent;
  \item improve visual descriptiveness and clarity;
  \item add appropriate details about attributes, scene, composition, style, and image quality;
  \item avoid introducing objects or meanings that are inconsistent with the original prompt;
  \item be suitable for T2I generation.
\end{itemize}

\textbf{Original prompt:}

\{x\}

\textbf{Output format:}

Optimized prompt: \textless Your rewritten prompt \textgreater
\end{tcolorbox}

\begin{tcolorbox}[
  promptbox,
  title={Visual Feedback Template for $\mathcal D_{\mathrm{fb}}$}
]
[IMAGE]

Please evaluate the generated image according to the given prompt.

\textbf{Prompt:}

\{p\}

\textbf{Task.}
Assess the image from three dimensions: semantic consistency, aesthetic quality, and preference alignment. For each dimension, provide a discrete score from 1 to 10 and a concise image-grounded explanation. The scoring should follow the scoring reference box provided below in Appendix ~\ref{box:scoring_reference}. The three dimensions should be judged independently.

\textbf{Output format:}

Semantic consistency score: \textless score from 1 to 10 \textgreater

Semantic consistency feedback: \textless concise explanation \textgreater

Aesthetic quality score: \textless score from 1 to 10 \textgreater

Aesthetic quality feedback: \textless concise explanation \textgreater

Preference alignment score: \textless score from 1 to 10 \textgreater

Preference alignment feedback: \textless concise explanation \textgreater

\end{tcolorbox}

\begin{tcolorbox}[
  promptbox,
  title={Visual Score Template for $\mathcal D_{\mathrm{score}}$}
]
[IMAGE]

Please score the generated image according to the given prompt.

\textbf{Prompt:}

\{p\}

\textbf{Task.}
Assign three independent scores from 1 to 10 for semantic consistency, aesthetic quality, and preference alignment, following the scoring reference box provided below in Appendix ~\ref{box:scoring_reference}. Only output the numerical scores without additional explanations.

\textbf{Output format:}

Semantic consistency score: \textless score from 1 to 10 \textgreater

Aesthetic quality score: \textless score from 1 to 10 \textgreater

Preference alignment score: \textless score from 1 to 10 \textgreater

\end{tcolorbox}

\begin{tcolorbox}[
  title={Feedback-Guided Refinement Template for $\mathcal D_{\mathrm{fgr}}$},
  width=\linewidth,
  colback=gray!5,
  colframe=gray!60,
  fonttitle=\bfseries,
  breakable
]
[IMAGE]

Please refine the original T2I prompt according to the generated image and the visual feedback.

\textbf{Original prompt:}

\{x\}

\textbf{Visual feedback:}

\{$\mathcal{F}^{*}_{x}$\}

\textbf{Task:}

Rewrite the original prompt to improve the next generated image. The refined prompt should:
\begin{itemize}
  \item preserve the original user intent;
  \item correct semantic omissions, weak attribute binding, and relationship-level inconsistencies mentioned in the feedback;
  \item improve aesthetic quality by adding appropriate visual details about composition, lighting, style, and image quality;
  \item improve preference alignment while avoiding unnecessary or inconsistent content;
  \item remain concise and suitable for T2I generation.
\end{itemize}

\textbf{Output format:}

Refined prompt: \textless Your feedback-guided optimized prompt \textgreater
\end{tcolorbox}

For $\mathcal D_{\mathrm{fb}}$ and $\mathcal D_{\mathrm{score}}$, all visual scores are assigned according to a shared 1--10 scoring rubric. This shared reference ensures consistent criteria between feedback annotation and reward score generation. The detailed scoring rubric is provided below.

\begin{tcolorbox}[
  promptbox,
  title={Scoring Reference for Visual Evaluation},
  label={box:scoring_reference}
]
All visual scores are assigned on a discrete scale from 1 to 10, where higher scores indicate better quality. The three dimensions are evaluated independently.

\textbf{Semantic consistency.}
Score 1: almost unrelated to the prompt.
Score 2: only a few prompt elements are captured.
Score 3: some relevant content is present but with clear semantic errors.
Score 4: the main intent is partially followed, with several missing or inconsistent details.
Score 5: the main idea is captured, but fine-grained attributes or relations are weak.
Score 6: most major elements are correct, with remaining detail-level errors.
Score 7: generally consistent with minor omissions.
Score 8: accurately reflects most objects, attributes, relations, and scene details.
Score 9: highly faithful with negligible imperfections.
Score 10: fully matches the prompt, including objects, attributes, relations, actions, and scene details.

\textbf{Aesthetic quality.}
Score 1: severe aesthetic flaws and poor visual balance.
Score 2: major issues in composition, color, lighting, or clarity.
Score 3: limited visual appeal with obvious flaws.
Score 4: understandable but weak in composition or harmony.
Score 5: acceptable but generic or noticeably improvable.
Score 6: moderately appealing but lacking refinement.
Score 7: good composition, color, lighting, and overall appeal.
Score 8: visually strong, coherent, and well balanced.
Score 9: excellent harmony, creativity, and visual impact.
Score 10: outstanding aesthetic quality with exceptional composition, balance, and creativity.

\textbf{Preference alignment.}
Score 1: highly unlikely to satisfy human preference.
Score 2: low appeal with major issues in naturalness, coherence, or style.
Score 3: partially acceptable but with obvious preference-related flaws.
Score 4: limited appeal and possibly awkward or inconsistent.
Score 5: acceptable but not especially attractive or satisfying.
Score 6: moderately preferred with reasonable coherence and appeal.
Score 7: likely to be preferred by users.
Score 8: strongly preferred, with appealing content and coherent style.
Score 9: highly preferred with strong attractiveness and satisfaction.
Score 10: exceptionally satisfying and highly aligned with human preference.
\end{tcolorbox}




\section{Additional Results on FLUX.2-klein}
\label{app:flux_results}
\begin{table*}[t]
  \centering
  \caption{Holistic image-level evaluation results on Flux.2-klein 4B. The best and second-best results in each setting are highlighted in bold and underlined, respectively.}
  \label{tab:flux_beautifulprompt_results}
  \small
  \setlength{\tabcolsep}{5.5pt}
  \renewcommand{\arraystretch}{1.12}
  \begin{tabular}{@{}lccccc@{}}
    \toprule
    Method & CLIPScore $\uparrow$ & Aesthetic $\uparrow$ & PickScore $\uparrow$ & HPSv2 $\uparrow$ & ImageReward $\uparrow$ \\
    \midrule
    
    Original Prompt
    & 0.285 & 6.129 & 20.782 & 0.274 & 1.066 \\
    \midrule

    \rowcolor{gray!12}
    \multicolumn{6}{@{}l}{\textit{Text-only prompt optimization}} \\
    BestPrompt
    & 0.274 & 6.215 & 20.576 & 0.283 & 1.096 \\
    BeautifulPrompt
    & 0.249 & 6.201 & 21.505 & 0.285 & 1.112 \\
    Promptist
    & \uline{0.287} & 6.127 & 20.802 & 0.279 & 1.147 \\
    NeuroPrompts
    & 0.265 & 6.180 & \uline{21.945} & 0.270 & \uline{1.198} \\
    Qwen3-VL-8B
    & 0.251 & 6.188 & 20.247 & 0.281 & 1.074 \\
    PRISM-SFT
    & 0.283 & \uline{6.224} & 21.584 & \uline{0.295} & 1.181\\
    PRISM
    & \textbf{0.292} & \textbf{6.296} & \textbf{22.319} & \textbf{0.301} & \textbf{1.205} \\
    \midrule

    \rowcolor{gray!12}
    \multicolumn{6}{@{}l}{\textit{Text-image feedback-based prompt optimization}} \\
    VisualPrompter
    & \textbf{0.308} & 6.124 & 21.286 & 0.286 & 0.981 \\
    TIR
    & 0.288 & 6.184 & 21.594 & 0.282 & 0.989 \\
    OPT2I
    & \uline{0.302} & 6.163 & 21.448 & 0.274 & 1.017 \\
    GPT-4o
    & 0.272 & \uline{6.228} & 20.946 & 0.296 & 1.218 \\
    Qwen3-VL-235B-A22B
    & 0.273 & 6.145 & \uline{22.659} & 0.291 & \uline{1.247} \\
    Qwen3-VL-8B
    & 0.258 & 6.118 & 21.702 & 0.287 & 1.193 \\
    PRISM-SFT
    & 0.276 & 6.209 & 22.371 & \uline{0.302} & 1.231 \\
    PRISM
    & 0.300 & \textbf{6.334} & \textbf{22.948} & \textbf{0.306} & \textbf{1.286} \\
    \bottomrule
  \end{tabular}
\end{table*}

We provide the comparison results on FLUX.2-klein 4B with 4 denoising steps as additional evidence of cross-generator generalization. The evaluation protocol follows the main experiments, with all methods tested under the same settings.

\textbf{Holistic image-level evaluation.}
Table~\ref{tab:flux_beautifulprompt_results} reports the results on the BeautifulPrompt test set. Consistent with the Stable Diffusion 3.5 results, our method achieves stronger overall performance across alignment, aesthetic quality, and preference-related metrics. This shows that PRISM can improve image-level generation quality on FLUX.2-klein as well.

\textbf{Fine-grained semantic alignment.}
Table~\ref{tab:flux_fine_grained_results} presents the results on T2I-CompBench and TIFA. PRISM also obtains better fine-grained semantic alignment, indicating that the optimized prompts better preserve compositional attributes, object relations, and detail-level prompt-image consistency.

\begin{figure*}[htbp]
  \includegraphics[width=1.0\linewidth]{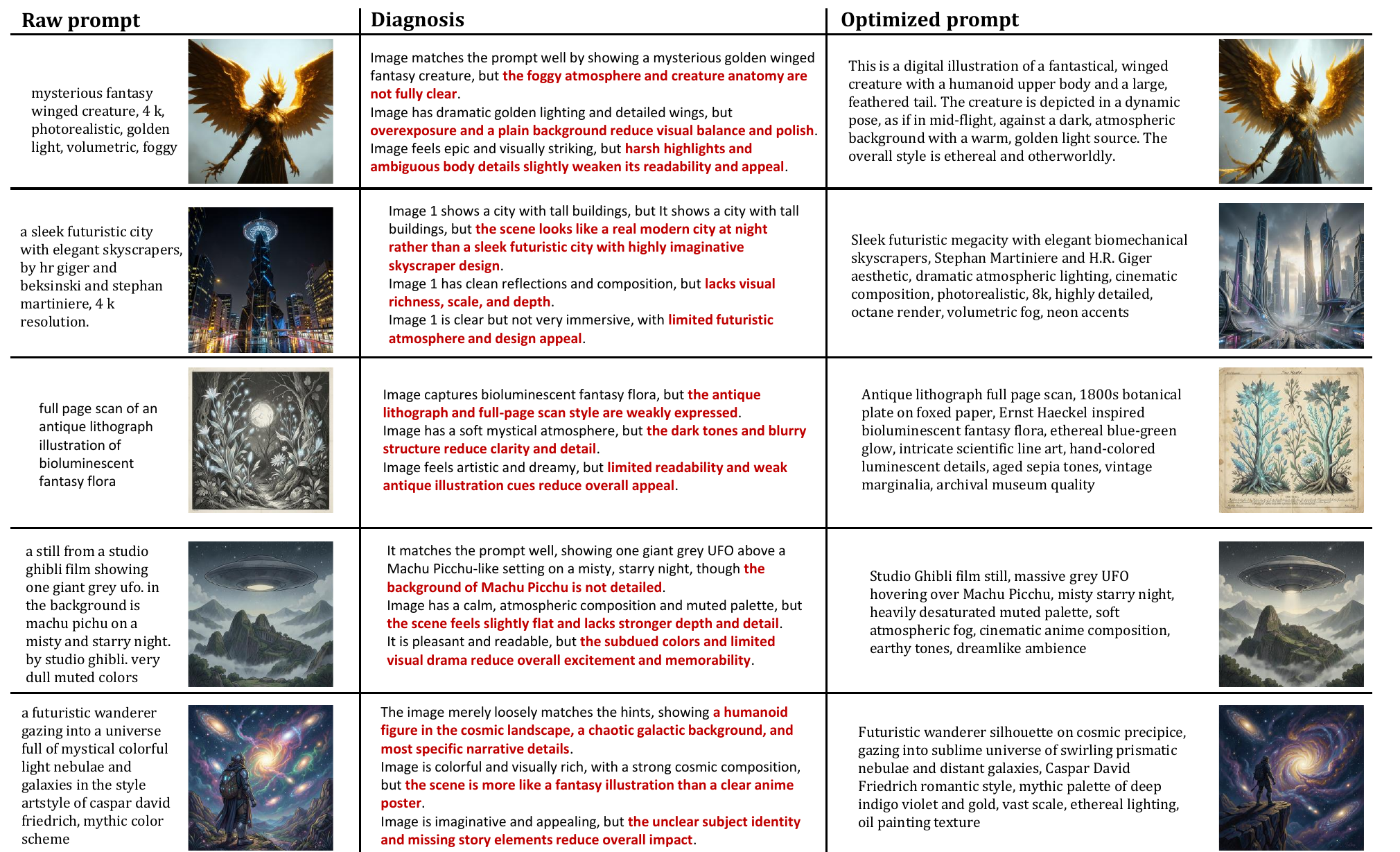}
  \caption{Qualitative examples of PRISM. 
  Each example shows the original prompt, initial generated image, structured visual feedback, optimized prompt, and final generated image, illustrating how diagnosed visual issues guide targeted prompt refinement.}
  \label{fig:qualitative_flux}
\end{figure*}

\textbf{Qualitative analysis.}
Figure~\ref{fig:qualitative_flux} shows qualitative comparisons on FLUX.2-klein. Compared with the original prompts and baseline prompt optimization methods, PRISM generates images with more faithful prompt-image correspondence and richer visual details. In particular, the optimized prompts help recover missing attributes, improve object-level relationships, and produce more coherent composition and style. These examples further show that PRISM can provide effective prompt corrections under a different T2I generator.
\begin{table*}[t]
  \centering
  \caption{Fine-grained semantic alignment evaluation on Flux.2-klein 4B. The best and second-best results are highlighted in bold and underlined, respectively.}
  \label{tab:flux_fine_grained_results}
  \small
  \setlength{\tabcolsep}{4.0pt}
  \renewcommand{\arraystretch}{1.12}
  \begin{tabular}{@{}lcccccccc@{}}
    \toprule
    \multirow{2}{*}{Method}
    & \multirow{2}{*}{TIFA $\uparrow$}
    & \multicolumn{7}{c}{T2I-CompBench $\uparrow$} \\
    \cmidrule(lr){3-9}
    & & Color & Shape & Texture & Spatial & NonSpatial & Complex & Avg. \\
    \midrule
    Original Prompt
    & 76.7 & 0.785 & 0.516 & 0.634 & 0.266 & 0.271 & 0.305 & 0.463 \\
    
    \midrule
    VisualPrompter
    & 88.2 & \uline{0.841} & \uline{0.602} & 0.701 & 0.410 & 0.322 & \uline{0.410} & \uline{0.548} \\
    TIR
    & 83.1 & 0.828 & 0.566 & 0.723 & 0.363 & 0.316 & 0.391 & 0.531 \\
    OPT2I
    & 85.7 & 0.825 & 0.599 & \uline{0.732} & 0.394 & 0.321 & 0.380 & 0.542 \\
    GPT-4o
    & \uline{90.8} & 0.813 & 0.555 & 0.721 & \textbf{0.436} & 0.315 & 0.386 & 0.538 \\
    Qwen3-VL-235B-A22B
    & 90.2 & 0.816 & 0.584 & 0.716 & 0.405 & \uline{0.326} & 0.402 & 0.541 \\
    Qwen3-VL-8B
    & 88.0 & 0.794 & 0.587 & 0.691 & 0.379 & 0.320 & 0.372 & 0.524 \\
    PRISM-SFT
    & 89.5 & 0.801 & 0.592 & 0.719 & 0.393 & 0.318 & 0.403 & 0.538 \\
    PRISM
    & \textbf{91.3} & \textbf{0.852} & \textbf{0.609} & \textbf{0.738} & \uline{0.425} & \textbf{0.330} & \textbf{0.414} & \textbf{0.561} \\
    \bottomrule
  \end{tabular}

  \vspace{1mm}
  \begin{minipage}{0.96\textwidth}
  \footnotesize
  \end{minipage}
\end{table*}

Overall, the FLUX.2-klein results are consistent with the main experiments, further supporting the robustness and cross-generator applicability of PRISM.

\section{Artifact Licenses, Terms of Use, and Intended Use}
\label{app:artifact_licenses_intended_use}

We use existing publicly available artifacts only for research purposes, including pretrained models, text-to-image generators, datasets, benchmarks, and evaluation tools. 
The main artifacts include Qwen3-VL, Stable Diffusion 3.5, FLUX.2-klein, BeautifulPrompt, DiffusionDB, JourneyDB, Pick-a-Pic, T2I-CompBench, TIFA, CLIPScore, ImageReward, PickScore, and HPSv2. 
We follow the licenses, terms of use, and access conditions specified by the original creators and distribution platforms. 
Our use of these artifacts is consistent with their intended research and evaluation purposes. 
The derived training and evaluation data constructed in this work are intended only for research on text-to-image prompt optimization and visual-feedback-based model analysis, and should not be used outside research contexts or for generating harmful, misleading, or unauthorized content. 
No existing artifact is redistributed as part of this submission.

\section{AI Assistance Statement}
\label{app:ai_assistance}

The authors used AI-assisted tools only for grammar checking, formatting inspection, and language polishing of the manuscript. 
No AI tools were used to generate the research ideas, formulate the method, design or conduct experiments, analyze results, create figures or tables, or draw scientific conclusions. 
All technical content, experimental results, claims, and final manuscript decisions were written, reviewed, and verified by the authors, who take full responsibility for the accuracy, originality, and integrity of the work.

\end{document}